%% file: main.tex
\documentclass{article}
\PassOptionsToPackage{numbers, compress}{natbib}

\usepackage[nonatbib,preprint]{main}
\usepackage[utf8]{inputenc}
\usepackage[T1]{fontenc}
\usepackage{hyperref}
\usepackage{url}
\usepackage{booktabs}
\usepackage{amsfonts}
\usepackage{nicefrac}
\usepackage{microtype}

\usepackage{multirow}
\usepackage{times}
\usepackage{epsfig}
\usepackage{amsmath} 
\usepackage{subfigure}
\usepackage{makecell}
\usepackage[dvipsnames]{xcolor}

\newcommand{\tablestyle}[2]{\setlength{\tabcolsep}{#1}\renewcommand{\arraystretch}{#2}\centering\footnotesize}
\newlength\savewidth
\usepackage[dvipsnames]{xcolor}

\usepackage[square,numbers,sort&compress]{natbib}
\hypersetup{
    colorlinks,
    citecolor=RoyalBlue,
}

\usepackage{tikz}
\usepackage{calc}

\usepackage{amsmath,amssymb,amsthm,array,amsfonts,mathtools}

\usepackage{algpseudocode}  
\usepackage{amsmath}
\usepackage{verbatim}

\DeclareMathOperator{\SE}{SE}

\DeclareMathOperator{\SO}{SO}

\input{math_commands.tex}

\title{Vidu4D: Single Generated Video to High-Fidelity 4D Reconstruction with Dynamic Gaussian Surfels}

\author{\hskip-0.15in Yikai Wang\thanks{Equal contribution. \quad $^\dagger$The corresponding author.}$^{\ \,1}$, Xinzhou Wang\footnotemark[1]$^{\ \,1,2,3}$, Zilong Chen$^{1,2}$, 
Zhengyi Wang$^{1,2}$, Fuchun Sun$^1$, Jun Zhu$^{\dagger1,2}$
\\$^1$Department of Computer Science and Technology, BNRist Center, Tsinghua University\\$^2$ShengShu \quad $^3$College of Electronic and Information Engineering, Tongji University\\ \texttt{\footnotesize yikaiw@outlook.com, wangxinzhou@tongji.edu.cn, dcszj@tsinghua.edu.cn}}

\begin{document}
\maketitle

\begin{abstract}

Video generative models are receiving particular attention given their ability to generate realistic and imaginative frames. Besides, these models are also observed to exhibit strong 3D consistency, significantly enhancing their potential to act as world simulators.  In this work, we present Vidu4D, a novel reconstruction model that excels in accurately reconstructing 4D (\emph{i.e.}, sequential 3D) representations from single generated videos, addressing challenges associated with non-rigidity and frame distortion. This capability is pivotal for creating high-fidelity virtual contents that maintain both spatial and temporal coherence. At the core of Vidu4D is our proposed \emph{Dynamic Gaussian Surfels} (DGS) technique. DGS optimizes time-varying warping functions to transform Gaussian surfels (surface elements) from a static state to a dynamically warped state. This transformation enables a precise depiction of motion and deformation over time. To preserve the structural integrity of surface-aligned Gaussian surfels, we design the warped-state geometric regularization based on continuous warping fields for estimating normals. Additionally, we learn refinements on rotation and scaling parameters of Gaussian surfels, which greatly alleviates texture flickering during the warping process and enhances the capture of fine-grained appearance details. Vidu4D also contains a novel initialization state that provides a proper start for the warping fields in DGS. Equipping Vidu4D with an existing video generative model, the overall framework demonstrates high-fidelity text-to-4D generation in both appearance and geometry. Project page: \linebreak\url{https://vidu4d-dgs.github.io}.

\end{abstract}

\section{Introduction}

The field of multimodal generation exhibits significant advancements and holds great promise for various applications. Recently, video generative models have garnered attention for their remarkable capability to craft immersive and lifelike frames~\cite{videoworldsimulators2024,bao2024vidu}. These models produce visually stunning content while also exhibiting strong 3D consistency~\cite{chen2024v3d,voleti2024sv3d}, largely increasing their potential to simulate realistic environments.

Parallel to these developments, high-quality 4D reconstruction has made great strides~\cite{pumarola2021d, TiNeuVox, DBLP:journals/tog/ParkSHBBGMS21, yang2023deformable3dgs, wu20234dgaussians}. This technique involves capturing and rendering detailed spatial and temporal information. When integrated with generative video technologies, 4D reconstruction potentially enables the creation of models that capture static scenes and dynamic sequences over time. This synthesis provides a more holistic representation of reality, which is crucial for applications such as virtual reality, scientific visualization, and embodied artificial intelligence.

\begin{figure}[!ht]
\centering  
\subfigure[Prompt: A portrait captures the dignified presence of an orange cat with striking blue eyes. The cat wears a single pearl earring. Her head tilts in contemplation, reminiscent of a Dutch cap.
]{\includegraphics[width=1\linewidth]{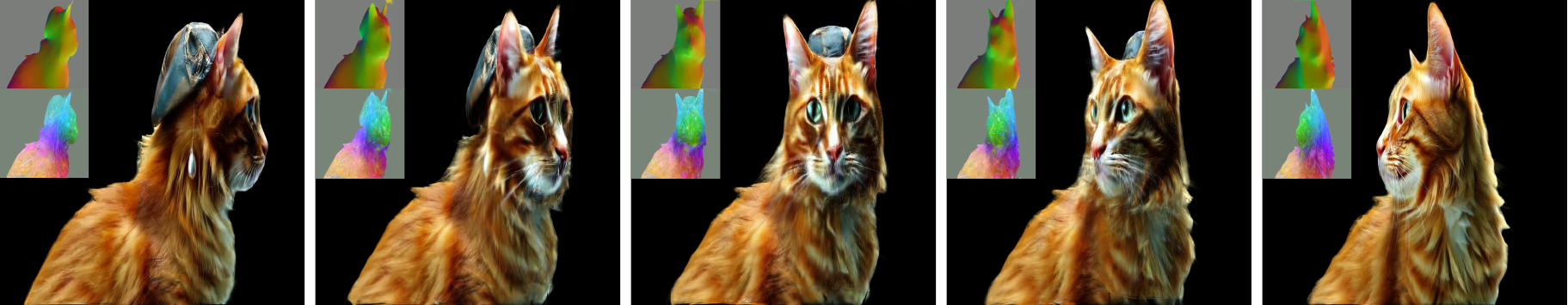}\label{fig:img_cat}}\\\vskip-0.03in
\subfigure[Prompt: A dragon with its hair blown by a strong wind. Devil enters the  soul with ethereal landscapes.]{\includegraphics[width=1\linewidth]{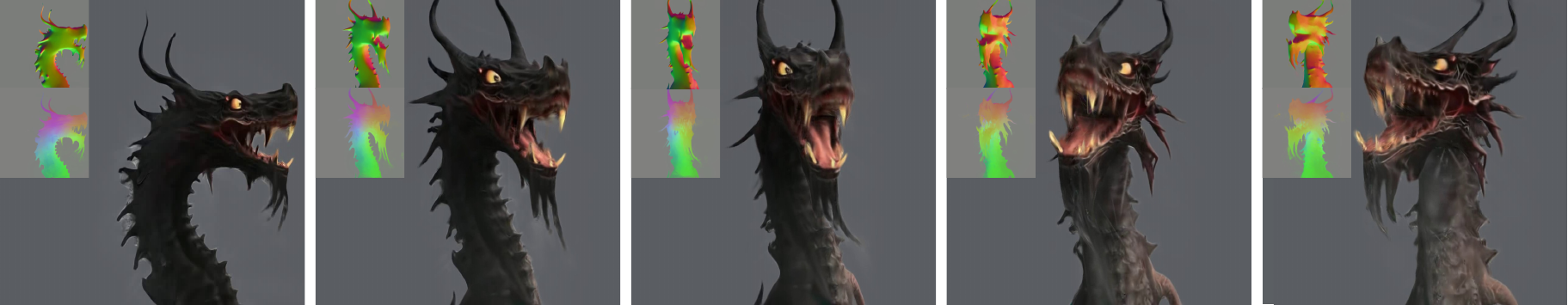}\label{fig:img_dragon}} \vskip-0.03in
\subfigure[Prompt: Light painting photo of a cheetah, cinematic.]{\includegraphics[width=1\linewidth]{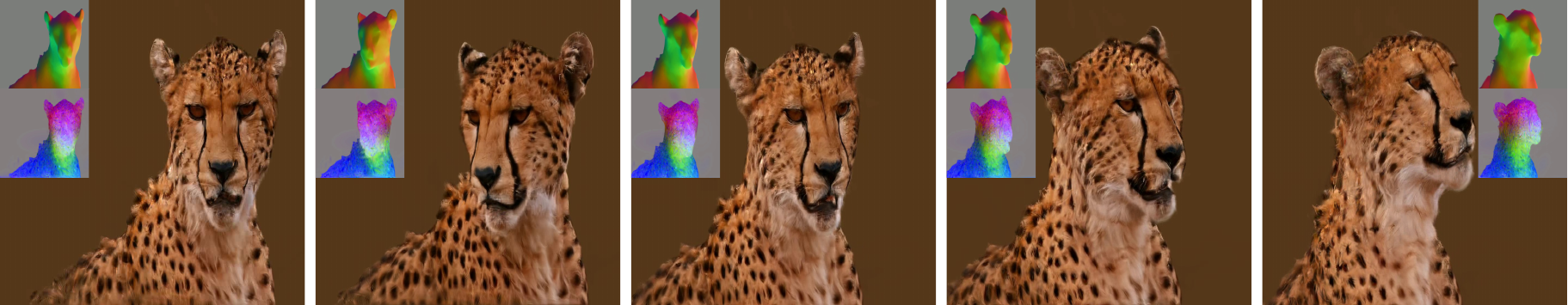}\label{fig:img_cheetah}} \vskip-0.03in
\subfigure[Prompt: A goldfish seemingly swimming through the air.]{\includegraphics[width=1\linewidth]{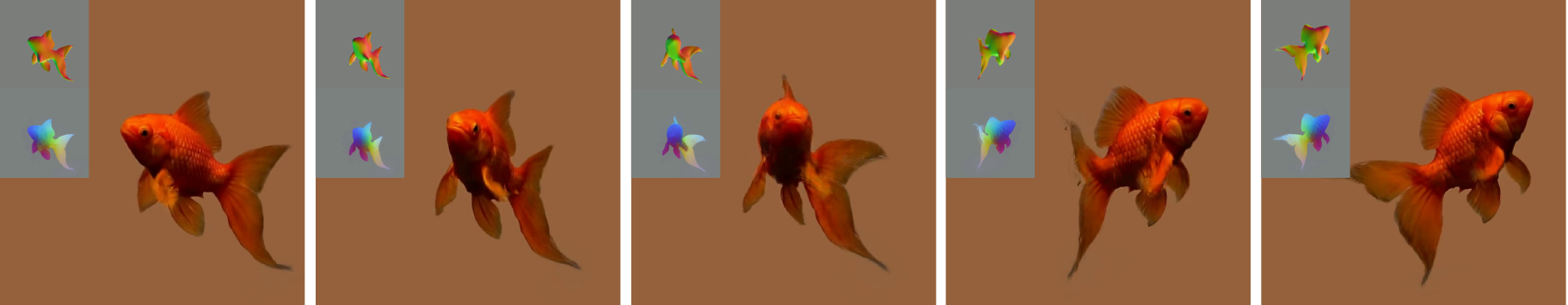}\label{fig:img_fish}} \vskip-0.03in
\subfigure[Prompt: A small, fluffy creature with an appearance reminiscent of a mythical being. The creature's fur texture is rendered in high detail. The monster's large eyes and open mouth express wonder and curiosity.]{\includegraphics[width=1\linewidth]{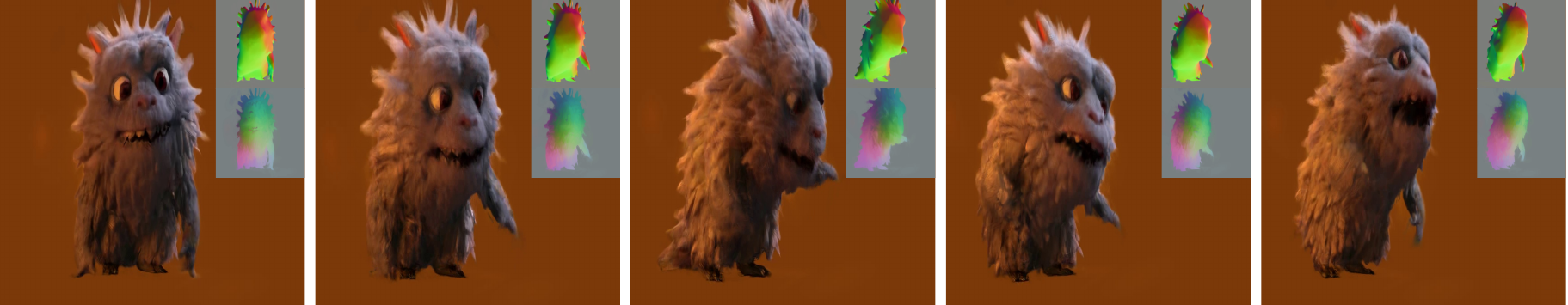}\label{fig:img_cheetah}} \vskip-0.03in
\subfigure[Prompt: An isolated coloured abstract sculpture with a dali shape.]{\includegraphics[width=1\linewidth]{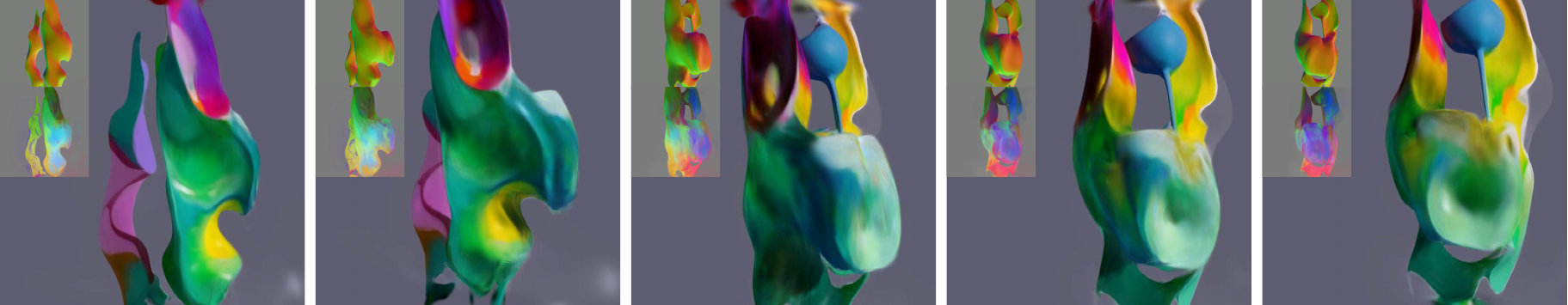}\label{fig:img_cheetah}}
\vskip-0.01in
\caption{{Text-(to-video)-to-4D samples generated by equipping Vidu4D with a pretrained video diffusion model~\cite{bao2024vidu}.} For each sample, we exhibit per-frame 3D rendering for novel-view color,  normal, and surfel feature. We observe that Vidu4D can reconstruct precisely detailed and photo-realistic 4D representation. See our accompanying videos in our \href{https://vidu4d-dgs.github.io}{project page}  for better visual quality.}
\vskip-0.09in
\label{fig:intro-image}
\end{figure}

However, achieving high-fidelity 4D reconstruction from generated videos poses great challenges. Non-rigidity and frame distortion are prevalent issues that can undermine the temporal and spatial coherence of the reconstructed content, thus complicating the creation of a seamless and coherent depiction of dynamic subjects.

In this work, we introduce Vidu4D, a novel reconstruction pipeline designed to accurately reconstruct 4D representations from single generated videos, facilitating the creation of  4D content with high precision in spatial and temporal coherence. Vidu4D contains two novel stages, namely, the initialization of non-rigid warping fields and Dynamic Gaussian Surfels (DGS), together enabling the reconstruction of high-fidelity 4D content with high-fidelity appearance and accurate geometry.

Specifically, the proposed DGS optimizes non-rigid warping functions that transform Gaussian surfels from static to dynamically warped states. This dynamic transformation accurately represents motion and deformation over time, crucial for capturing realistic 4D representations. Besides, DGS demonstrates superior 4D reconstruction performance due to two other key aspects. Firstly, in terms of geometry, DGS adheres to Gaussian surfels principles~\cite{Huang2DGS2024,Dai2024GaussianSurfels} to achieve precise geometric representation. Unlike existing methods, DGS incorporates warped-state normal consistency regularization to align surfels with actual surfaces with learnable continuous fields (\emph{w.r.t.} spatial coordinate and time) to ensure smooth warping when estimating normals. Secondly, for appearance, DGS learns additional refinements on the rotation and scaling parameters of Gaussian surfels by a dual branch structure. This refinement reduces the flickering artifacts during warping and allows for the precise rendering of appearance details, resulting in high-quality reconstructed 4D representations. 

By integrating Vidu4D with an existing powerful video generative model named Vidu~\cite{bao2024vidu}, the overall framework demonstrates exceptional capabilities in text-to-4D generation. We provide 4D visualization results in  Fig.~\ref{fig:intro-image}. Extensive experiments based on the generated videos verify the effectiveness of our method compared to current state-of-the-art methods.

\section{Related works}

\textbf{3D representation.} Transforming 2D images into 3D representations has long been a central challenge in the field. Initially, triangle meshes were favored for their compactness and compatibility with rendering pipelines \cite{buehler2001unstructured, debevec1996modeling, waechter2014let, wood2000surface, riegler2020free, thies2019deferred}. However, the transition to more sophisticated volumetric methods was inevitable due to the limitations of surface-based approaches. Early volumetric representations included voxel grids \cite{sitzmann2019deepvoxels, lombardi2019neural, penner2017soft, kutulakos2000theory} and multi-plane images \cite{zhou2018stereo, flynn2019deepview, mildenhall2019local, srinivasan2019pushing, srinivasan2020lighthouse, tucker2020single}, which, despite their straightforwardness, demanded intricate optimization strategies. The introduction of neural radiance fields (NeRF) \cite{DBLP:conf/eccv/MildenhallSTBRN20} marked a significant advancement, offering an implicit volumetric neural representation that could store and query the density and color of each point, leading to highly realistic reconstructions. The NeRF paradigm has since been improved upon in terms of reconstruction quality \cite{barron2021mip, barron2023zipnerf, kerbl3Dgaussians, ma2021deblur, wang20224k} and rendering~\cite{reiser2023merf, hedman2021snerg, yu2021plenoctrees, reiser2021kilonerf, liu2020neural, rebain2021derf, lindell2021autoint, garbin2021fastnerf, Hu_2022_CVPR, cao2022mobiler2l, wang2022r2l, lombardi2021mixture}.  To address the limitations of NeRF, such as rendering speed and memory usage, recent work dubbed 3D Gaussian splatting (3DGS)~\cite{kerbl3Dgaussians} has proposed anisotropic Gaussian representations with GPU-optimized tile-based rasterization. This has opened up new avenues for surface extraction \cite{Huang2DGS2024, guedon2023sugar}, generation \cite{chen2023text, tang2024lgm, xu2024grm}, and large-scale scene reconstruction \cite{liu2024citygaussian, shuai2024LoG, hierarchicalgaussians24}, with 3DGS emerging as a universal representation for 3D scenes and objects. Gaussian surfels methods~\cite{Huang2DGS2024,Dai2024GaussianSurfels} further exhibit advantages in modeling accurate geometry. While these methods have significantly advanced the field of static 3D representation, capturing the dynamic aspects of real-world scenes with non-rigid motion and deformation introduces a distinct set of challenges that demand innovative solutions.

\textbf{Dynamic reconstruction and generation.} The dynamic reconstruction of scenes from video captures presents a more complex challenge than static reconstruction, necessitating the capture of non-rigid motion and deformation over time \cite{li2022tava, peng2021neural, su2021anerf, zhang2021stnerf, wang2023animatabledreamer}. Traditional methods have explored dynamic reconstruction using synchronized multi-view videos \cite{lombardi2019neural, li2022streaming, wang2023rpd, attal2023hyperreel, song2022nerfplayer, cao2023hexplane, wang2022mixed, wang2022fourier, bansal20204d, peng2023representing, wang2023masked} or have focused on specific dynamic elements like humans or animals. More recently, there has been a shift towards reconstructing non-rigid objects from monocular videos, which is a more practical yet challenging scenario. One approach involves incorporating time as an additional input to the neural radiance field \cite{li2022neural, kplanes_2023, cao2023hexplane,DBLP:conf/cvpr/YangVNRVJ22}, allowing for explicit querying of spatiotemporal information. Another line of research decomposes the spatiotemporal radiance field into a canonical space and a deformation field, representing spatial attributes and their temporal variations \cite{pumarola2021d, TiNeuVox, DBLP:journals/tog/ParkSHBBGMS21, dycheck, park2021nerfies, TiNeuVox, shao2023tensor4d, liu2023robust, liu2022devrf, Zhao_2022_CVPR, tretschk2021non, jiang2022alignerf, du2021nerflow, gao2021dynamic, li2020neural, xian2021space}. With advancements in 3DGS, deformable-GS \cite{yang2023deformable3dgs} and 4DGS \cite{wu20234dgaussians} have been developed, utilizing neural deformation fields with multi-layer perception (MLP) and triplane, respectively. SCGS~\cite{huang2023sc} and dynamic 3D Gaussians \cite{luiten2023dynamic} also advance the field by modeling time-varying scenes. Building on these advances, our work introduces dynamic Gaussian surfels, a novel extension of Gaussian representations that enhances the quality of both appearance and surface reconstruction under dynamic scenarios. In the realm of 3D or 4D generation, our approach diverges from recent progress in optimization-based \cite{poole2022dreamfusion, wang2023prolificdreamer, lin2023magic3d, chen2023fantasia3d, chen2023text,wang2023animatabledreamer,singer2023text,ling2023align,bah20244dfy}, feed-forward \cite{hong2023lrm, zou2023triplane, wang2024crm}, and multi-view reconstruction methods \cite{chen2024v3d, long2023wonder3d, lu2023direct2} by leveraging a video generative model to achieve generation capabilities. Our primary focus is on preserving high-quality appearance and geometrical integrity from generated videos. This results in a generation process that not only captures the nuances of motion and deformation but also maintains the high standards of realism and detail that are essential for creating immersive and lifelike virtual 3D representations.

\section{Method}
In this section, we first introduce the basic problem definition of 4D reconstruction (see Sec.~\ref{subsec:defination}).  We then present our method dubbed Dynamic Gaussian Surfels (DGS) for accurately modeling both the appearance and geometry during the 4D reconstruction with large non-rigidity (see Sec.~\ref{subsec:modeling}). Finally, we introduce Vidu4D as a reconstruction pipeline and the overall framework for performing a generation task (see Sec.~\ref{subsec:vidu4d}).

\subsection{Problem Definition}
\label{subsec:defination} 

When given a single sequence of RGB video with $T$ frames, the goal of 4D reconstruction is to determine a sequential 3D representation that could be rendered to fit each video frame as much as possible. Specifically, suppose the 3D representation for the $t$-th frame (termed as time $t$) is parameterized by $\theta_t$, where $t=1,\cdots,T$.  Given a differentiable rendering mapping  $\vg$, we could obtain the rendered color at the frame pixel $\bar\bx^t\in\mathbb{R}^2$. We choose volume rendering as commonly adopted in NeRF~\cite{DBLP:conf/eccv/MildenhallSTBRN20}, Gaussian Splatting~\cite{kerbl3Dgaussians}, and Gaussian Surfels~\cite{Huang2DGS2024,Dai2024GaussianSurfels}. The optimization of 4D reconstruction can be implemented by minimizing the empirical loss as 
\begin{equation}
\label{eq:loss}
    \min_{\theta}\frac{1}{T}\sum_{t=1}^{T}\sum_{\bar\bx^t}\mathcal{L}\Big(\bc(\bar\bx^t)=\vg\big(\theta_t, \{\bx^t_i\}_{i=1,\cdots,N}\big),  \hat\bc(\bar\bx^t)\Big),
\end{equation}
where $\bx^t_i \in \mathbb{R}^3$ is the $i$-th 3D point sampled or intersected with Gaussian primitives along the ray that emanates from the frame pixel $\bar\bx^t$; $N$ is the number of sampled or intersected points per ray; $\bc(\bar\bx^t)$ and $\hat\bc(\bar\bx^t)$ are the rendered color and the observed color at  $\bar\bx^t$, respectively.

\subsection{Dynamic Gaussian Surfels}
\label{subsec:modeling} 

By optimizing Eq.~(\ref{eq:loss}), essentially our goal is to build a sequential  3D representation that could deform to be consistent with each 2D frame. We first start by considering an ideal video exhibiting different views of the same static object without object deformation, movement, or video distortion. To model the 3D  representation with high appearance fidelity and geometry accuracy, we follow the method of using differentiable 2D Gaussian primitives as proposed by recent Gaussian Surfels advances~\cite{Huang2DGS2024,Dai2024GaussianSurfels}. Specifically, the  $k$-th Gaussian surfel (of the total $K$)  is characterized by a central point $\bp_k^*\in\mathbb{R}^3$ and a local coordinate system centered at $\bp_k^*$ with  two principal tangential vectors $\bt_u^*\in\mathbb{R}^{3\times 1}$, $\bt_v^*\in\mathbb{R}^{3\times 1}$ and scaling factors $s_u^*\in\mathbb{R}$, $s_v^*\in\mathbb{R}$. Here, we use the notation ``$*$'' to represent parameters in the static state. A Gaussian surfel is computed as a 2D Gaussian defined in a local tangent plane in the world space. Following~\cite{Huang2DGS2024}, for any point $\bu=(u,v)$ located on the  $uv$ coordinate system centered at $\bp_k^*$, its coordinate  in the world space, denoted as $P_k^*(\bu)\in\mathbb{R}^{3\times 1}$, is computed by
\begin{align}
P_k^*(\bu)=\bp_k^*+s_u^*\bt_u^*u+s_v^*\bt_v^*v=\begin{bmatrix}
        \bR_k^*\bS_k^* & \bp_k^* 
    \end{bmatrix}(u,v,1,1)^\top,
\label{eq:2dgaussian}
\end{align}
where  $\bR_k^*=[\bt_u^*,\bt_v^*, \bt_u^*\times \bt_v^*]\in\SO(3)$ denotes the rotation matrix, and the diagonal matrix $\bS_k^*=\mathrm{diag}(s_u^*, s_v^*, 0)\in\mathbb{R}^{3\times 3}$ denotes the scaling  matrix.

In this work, our focus is on 4D reconstruction from a single generated video, which may exhibit significant non-rigidity, distortion, or illumination changes. We introduce \textbf{Dynamic Gaussian Surfels (DGS)}, a method designed to achieve precise 4D reconstruction while accommodating non-rigidity and other time-varying effects.

Motivated by recent advancements in non-rigid reconstruction methods~\cite{park2021nerfies,DBLP:conf/cvpr/YangVNRVJ22,wang2023animatabledreamer},  we aim to ensure that the target object maintains a consistent static state across different frames, thereby mitigating non-rigidity and distortion effects. To achieve this, we employ warping techniques on each Gaussian surfel represented by  $P_k^*(\bu)$, transforming them into a corresponding Gaussian surfel $P_k^t(\bu)$ at time  $t$, which is centered at $\bp^t_k\in\mathbb{R}^3$ with a rotation matrix $\bR_k^t\in\SO(3)$ and a scaling matrix $\bS_k^t\in\mathbb{R}^{3\times 3}$.

\textbf{Non-rigid warping for Gaussian surfels.} 
We now build the warping process from the static state to the warped state. We define a time-varying non-rigid warping function by leveraging $B$  bones as key points to ease the training of deformation.  In the static state, the $b$-th bone is represented by 3D Gaussian ellipsoids~\cite{DBLP:conf/cvpr/YangSJVCCRFL21} with the center ${\bc}^*_b\in\mathbb{R}^{3 \times 1}$, rotation matrix $\bV_b^*\in\mathbb{R}^{3\times3}$, and  diagonal scaling matrix $\boldsymbol{\Lambda}_b^*\in\mathbb{R}^{3\times3}$.  We let $\bJ_b^{t}\in\SE(3)$ represent a rigid transformation that moves the $b$-th bone from its static state to the warped state at time $t$.  For a 3D point $P_k^*(\bu)$, the skinning weight vectors $\bw^t\in\mathbb{R}^{B\times1}$  at time $t$ is calculated by the normalized Mahalanobis distance following~\cite{DBLP:conf/cvpr/YangVNRVJ22}
\begin{equation}
    {m}^t_b = \big(P_k^*(\bu)-{\bc}^t_b\big)^\top{\bf Q}^t_b\big((P_k^*(\bu)-{\bc}^t_b\big),\quad\bw^t=\sigma_\mathrm{softmax}\big({m}^t_1, {m}^t_2, \cdots,{m}^t_B\big)^\top,
 \label{eq:bone}
\end{equation}
where ${m}^t_b$ denotes the squared distance between $P_k^*(\bu)$ and the $b$-th bone; ${\bc}^t_b\in\mathbb{R}^{3 \times 1}$ is the center of the $b$-th bone at time $t$,  and ${\bf Q}^t_b = {\bV_b^t}^\top\boldsymbol{\Lambda}_b^*\bV_b^t$ is the precision matrix composed by the bone orientation matrix $\bV_b^t\in\mathbb{R}^{3\times3}$ at time $t$ and $\boldsymbol{\Lambda}_b^*$. Specifically, there is $(\bV_b^t|{\bc}^t)=\bJ_b^{t}(\bV_b^*|{\bc}^*)$ with ${\bc}^*_b$, $\bV_b^*$, and $\boldsymbol{\Lambda}_b^*$ being learnable parameters. $\sigma_\mathrm{softmax}$ is the $\mathrm{softmax}$ function.

In effect, $\mathbf{J}^t_{b}$ is achieved by non-linear mappings using a multi-layer perception (MLP) with $\SE(3)$ guaranteed, as will be given later in Eq.~(\ref{eq:nerf_bone}). The non-rigid warping function can be written as the weighted combination of $\bJ_b^{t}\in\SE(3)$, where we apply dual quaternion blend skinning (DQB)~\cite{DBLP:conf/si3d/KavanCZO07} to ensure valid $\SE(3)$ after combination,
\begin{align}
\bJ^{t}=\mathcal{R}\Big(\sum_{b=1}^{B} w_{b}^{t} \mathcal{Q}(\mathbf{J}^t_{b})\Big),
\label{eq:warpingJ}
\end{align}
where $w_{b}^{t}$ is the $b$-th element of $\bw^t$; $\mathcal{Q}$ and $\mathcal{R}$ denote the quaternion process and the inverse quaternion process, respectively. In this case, $\bJ^{t}\in\SE(3)$.

\begin{figure}[t]
    \centering
     \hskip-0.05in
\includegraphics[width=1\columnwidth]{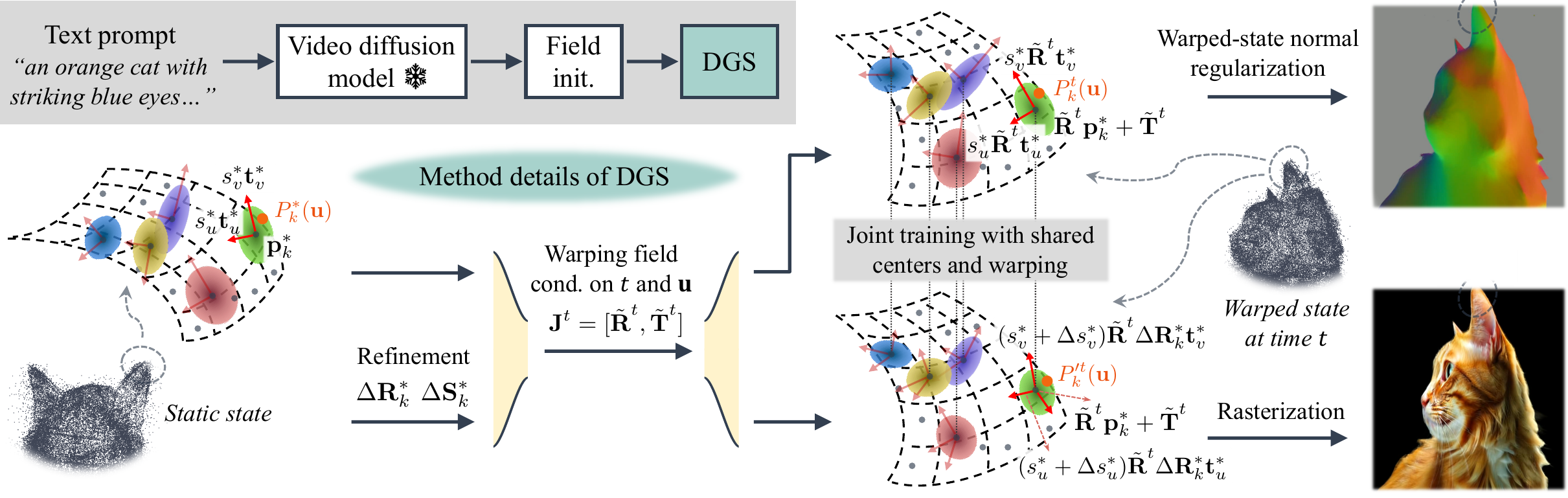}
    \vskip-0.01in
\caption{{Illustration of the overall framework and our DGS in detail.} For DGS, Gaussian surfels in the static state are transformed to the warped state by learning non-rigid warping functions conditioned on time $t$ and coordinate $\bu$. We incorporate warped-state normal regularization for accurate geometry, and refined rotation and scaling matrices of Gaussian surfels for detailed appearance. Both branches in the warped state, including with and without refinement, share the same centers of Gaussian surfels and the same warping functions. ``Field init.'' stands for field initialization as introduced in Sec.~\ref{subsec:vidu4d}.} 
\label{fig:method}
\end{figure}

We therefore rewrite the warping  as $\bJ^{t}=[
\tilde\bR^t,\tilde\bT^t] $ with the rotation $\tilde\bR^t\in\SO(3)$ and translation $\tilde\bT^t\in\mathbb{R}^{3}$, and apply the corresponding transformation to Eq.~(\ref{eq:2dgaussian}) by 
\begin{align}
P_k^t(\bu)=\bJ^{t}P_k^*(\bu)=\begin{bmatrix}
        \tilde\bR^t\bR_k^*\bS_k^* & \tilde\bR^t\bp_k^*+\tilde\bT^t \\
    \end{bmatrix}(u,v,1,1)^\top.
    \label{eq:warping}
\end{align}

Note that Eq.~(\ref{eq:warping}) holds for any given point $P_k^*(\bu)$ including the  center point of the $k$-th Gaussian surfel (\emph{i.e.}, $\bp_k^*$) when $\bu=(0,0)$. By deriving  Eq.~(\ref{eq:warping}), we enable connection of the warping function \emph{w.r.t.} to any point $\bu=(u,v)$ on the local coordinate system centered at $\bp_k^*$, which is needed later in Eq.~(\ref{eq:2dgs}) where $\bu$ is an intersection with Gaussian surfels and a ray that emanates from the frame pixel.

\textbf{Warped-state normal regularization.} 
To accurately capture the geometric representation, we follow similar methods in Gaussian Surfels~\cite{Huang2DGS2024,Dai2024GaussianSurfels} to add normal consistency regularization which encourages all Gaussian surfels to be locally aligned with the actual surfaces. Differently, unlike 3D reconstruction for static scenes, 4D reconstruction commonly faces non-rigidity and distortion. Thus simply performing regularization to promote surface-aligned Gaussian surfels like previous methods harms the structural integrity due to the non-rigid warping.

We therefore design a warped-state normal regularization. As mentioned,  each point $P_k^t(\bu)$ in the warped state at time $t$ is transformed from its corresponding static point $P_k^*(\bu)$ based on the warping function in Eq.~(\ref{eq:warping}), namely, $P_k^t(\bu)=\bJ^{t}P_k^*(\bu)$ with $\bJ^{t}$ composed by  $\bJ_b^{t}$. To maintain the structural integrity to a large extent when regularizing normal, we design $\bJ_b^{t}$ as a continuous field that takes both the point $P_k^*(\bu)$ (or equivalently, $\bu$ in the local coordinate system) and the time $t$ as conditions. By this setting, $\bJ_b^{t}$ is expected to change continuously with the change of $\bu$ or $t$. We implement the continuous field by using a NeRF-style MLP which directly outputs a 6-dimensional dual quaternion, and rely on  the inverse quaternion process $\mathcal{R}$ to guarantee $\SE(3)$, \emph{i.e.}, 
\begin{equation}
{\bf J}^t_b = \mathcal{R}\big(\mathbf{MLP}(\boldsymbol\gamma_b^t; \bu, t)\big),
\label{eq:nerf_bone}
\end{equation}
where $\boldsymbol\gamma_b^t$ is a learnable latent code for encoding the $b$-th bone at time $t$; both $\bu$ and $t$ are sent to the MLP as conditions to obtain ${\bf J}^t_b$. Thus $\bJ^{t}$ is also expected to be continuous \emph{w.r.t.} $\bu$ and $t$.

Based on the above design, the normal consistency loss at time $t$ is obtained similar to~\cite{Huang2DGS2024}, 
\begin{equation}
\mathcal{L}_{n} = \sum_{k=1}^K \omega_k (1-\bn_k^\top\bN^t),\quad \mathbf{N}^t(x,y) = \frac{\nabla_x \bp^t \times \nabla_y \bp^t}{|\nabla_x \bp^t \times \nabla_y \bp^t|},
\end{equation}
where $k$ indexes over intersected surfels along the ray that emanates from the frame pixel $\bar\bx$; $\omega_k = \alpha_k\,{\cG}_k(\bu(\bar\bx))\prod_{j=1}^{k-1} (1 - \alpha_j\,{\cG}_j(\bu(\bar\bx)))$ denotes the blending weight of the intersection point; $\bn_k$ represents the normal of the surfel that is oriented towards the camera; $\bN^t$, computed with finite differences, is the surface normal estimated by the nearby depth point $\bp^t$ at warped state time $t$. 

In summary, by learning a continuous warping field and aligning the surfel normal with the estimated surface normal in the warped state, we ensure that all Gaussian surfels locally approximate the actual object surface without being noticeably impaired by the non-rigid warping.

\textbf{Dual branch structure with refinement.}
 To further achieve fine-grained appearance and reduce the texture flickering during warping, we propose to learn refinement terms for adjusting the rotation matrices $\bR_k^{*}$ and scaling matrices $\bS_k^{*}$ (defined in Eq.~(\ref{eq:2dgaussian})) in the static state. 
We suppose the refinement terms  are $\Delta\bR_k^{*}\in\SO(3)$ and $\Delta\bS_k^{*}\in\mathbb{R}^{3\times 3}$, respectively. Note that the third-axis of $\Delta\bS_k^{*}$ is no longer necessarily $0$. During refinement, we remain the center points $\bp_k^*$ and the warping  $\bJ^{t}$ (\emph{i.e.}, including both $\tilde\bR^t$ and $\tilde\bT^t$) to be unchanged. The new warped process is formulated as,
\begin{align}
P_k^{\prime t}(\bu)=\begin{bmatrix}
        \tilde\bR^t(\Delta\bR_k^{*}\bR_k^{*})(\bS_k^{*}+\Delta\bS_k^{*}) & \tilde\bR^t\bp_k^*+\tilde\bT^t \\
    \end{bmatrix}(u,v,1,1)^\top.
    \label{eq:warping_refine}
\end{align}

During the training of DGS, we maintain two branches including one with refinement and one without. In the warped state, both branches are jointly trained with shared warping functions and centers of Gaussian primitives\footnote{Here, since the third-axis of the refined scaling matrix is not necessarily 0,  we adopt ``Gaussian primitive'' for commonly referring to both Gaussian surfel and the refined Gaussian.}. Due to the involvement of $\Delta\bR_k^{*}$ and $\Delta\bS_k^{*}$, both branches have different rotation and scaling matrices of Gaussian primitives.

\textbf{Rasterization.} Given a frame pixel $\bar\bx$ and a camera ray that emanates from $\bar\bx$, following the static-state methods to calculate intersection coordinates with Gaussian primitives along the ray~\cite{kerbl3Dgaussians,Huang2DGS2024}, we could obtain warped-state intersection coordinates based on Eq.~(\ref{eq:warping}) and  Eq.~(\ref{eq:warping_refine}). We then perform the volume rendering process~\cite{Huang2DGS2024} that integrates alpha-weighted appearance along the  ray  by 
\begin{equation}
\bc(\bar\bx) = \sum_{k} \bc_k\,\alpha_k\,\cG_k\big(\bu(\bar\bx)\big) \prod_{j=1}^{k-1} \big(1 - \alpha_j\,\cG_j\big(\bu(\bar\bx)\big)\big),
\label{eq:2dgs}
\end{equation}
where  $k$ indexes over intersected Gaussian primitives along the ray that emanates from the frame pixel $\bar\bx$; $\alpha_k$ and $\bc_k$ denote the opacity and view-dependent appearance parameterized with spherical harmonics of the $k$-th Gaussian surfel, respectively; $\cG_k(\bu(\bar\bx)) = \exp\left(-\frac{u^2+v^2}{2}\right)$  corresponds to the $k$-th intersection point $\bu(\bar\bx)$ which could be directly calculated when given $P_k^t(\bu)$ or $P_k^{\prime t}(\bu)$ and the corresponding local coordinate system. During implementation, $\cG_k(\bu(\bar\bx)))$ is further applied a low-pass filter following~\cite{botsch2005high,Huang2DGS2024}.

\begin{figure}[t]
    \centering
     \hskip-0.05in
\includegraphics[width=1\columnwidth]{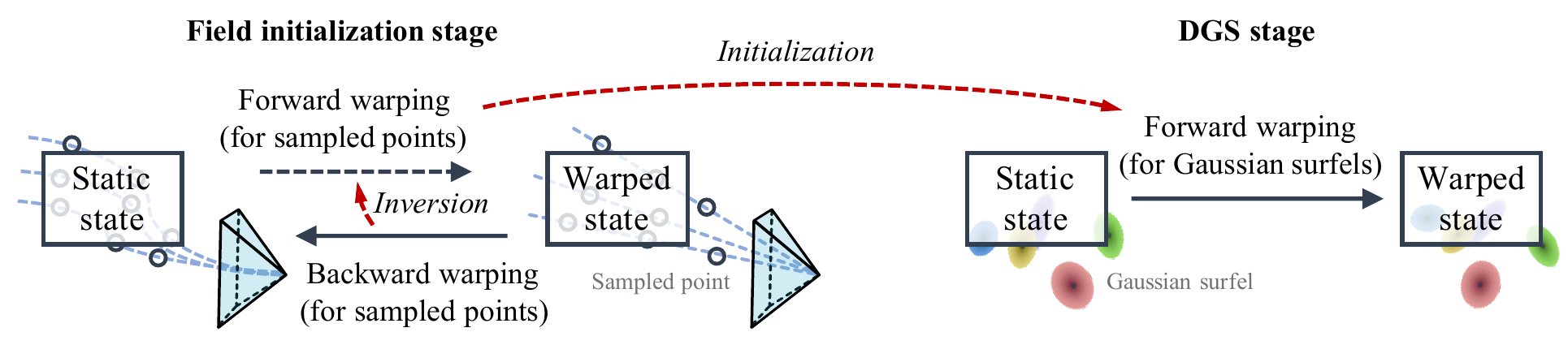}
    \vskip-0.01in
\caption{Illustration of the pipeline of Vidu4D, including the initialization stage and the DGS stage. }   \label{fig:vidu4d}\vskip-0.04in
\end{figure}

A detailed architecture of DGS is depicted in Fig.~\ref{fig:method}. Important symbols are summarized in Table~\ref{table:symbol}.

\subsection{Vidu4D}
\label{subsec:vidu4d}
Given that the camera trajectory of generated videos is unknown, SfM methods like COLMAP struggle to converge due to rigidity violations. Additionally, since the background of generated videos appears to exhibit soft deformation or flickering colors, proper estimation of camera/body poses through background SfM is hindered. These challenges often result in very few successful registrations, as demonstrated in previous monocular 4D reconstruction tasks~\cite{DBLP:conf/cvpr/YangVNRVJ22}.

In this part, we arrive at Vidu4D, a reconstruction pipeline comprising two key stages as illustrated in Fig.~\ref{fig:vidu4d}, including a field initialization stage and the DGS stage.  Specifically, we propose the field initialization as another key component of our pipeline to initialize the field in Eq.~(\ref{eq:nerf_bone}) of DGS for fast and stable convergence. We first train a neural SDF~\cite{DBLP:conf/nips/WangLLTKW21} using the same bone-based warping structure as utilized in our DGS. Unlike DGS, which warps Gaussian surfels from the static state to the warped state for rasterization, the neural SDF warps sampled points on camera rays from the warped state back to the static state. For the neural SDF part, we optimize the backward warping and learn a forward warping as the inversion of the backward warping by employing a cycle loss, inspired by~\cite{DBLP:journals/corr/abs-2206-15258,DBLP:conf/cvpr/YangVNRVJ22}. We then initialize the MLP to obtain warping functions ${\bf J}^t_b$ by the MLP learned by the neural SDF part. We provide more details in our Appendix.

With our field initialization before  DGS, our Vidu4D is capable of performing a text-(to-video)-to-4D generation task with the integration of existing video diffusion models.

\begin{figure}[t]
    \centering  \hskip-0.06in
\includegraphics[width=1\columnwidth]{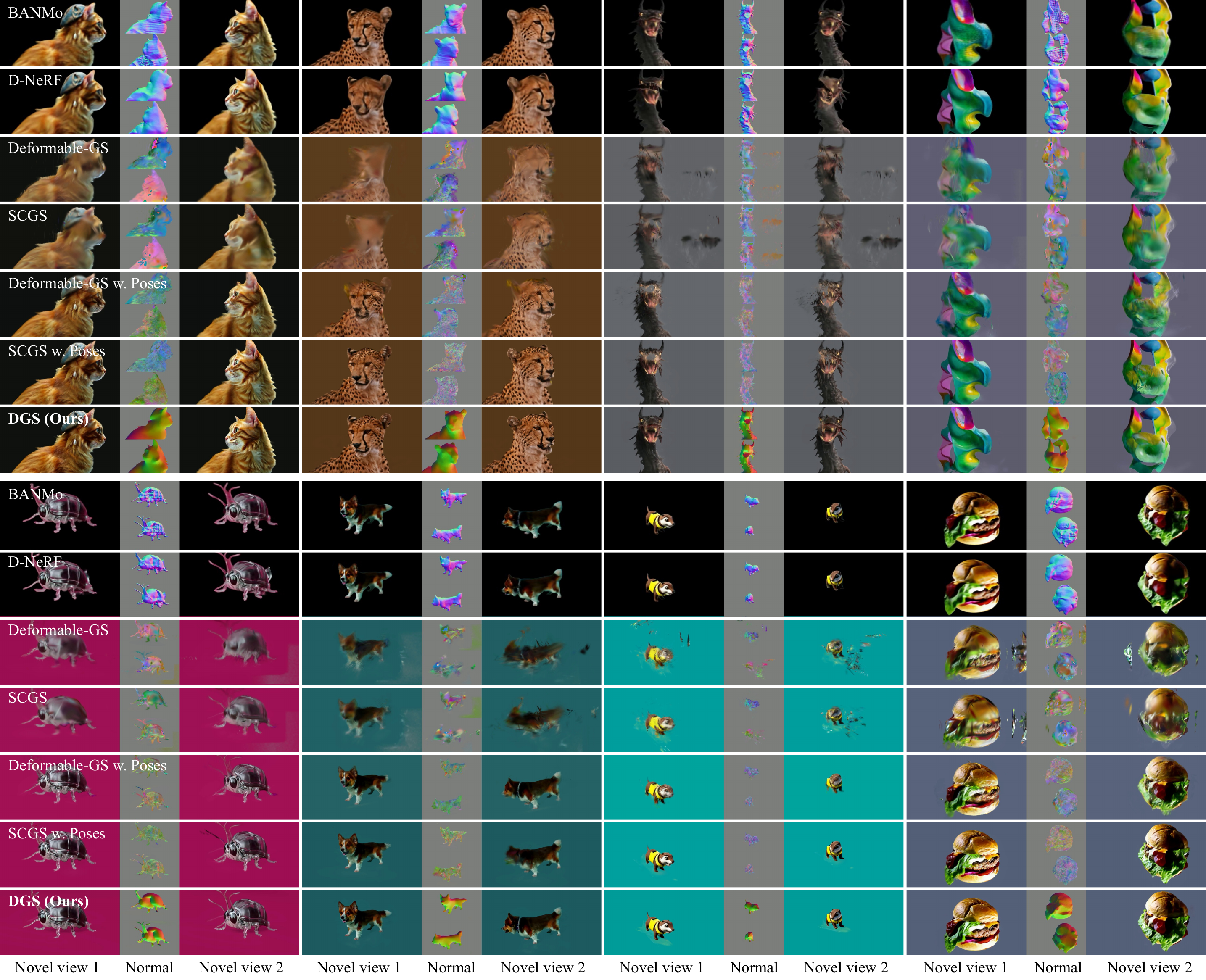}
    \vskip-0.03in
    \caption{Novel-view qualitative evaluation compared with SOTA methods including NeRF-based methods (BANMo~\cite{DBLP:conf/cvpr/YangVNRVJ22} and D-NeRF~\cite{pumarola2021d}) and Gaussian splatting-based methods (Deformable-GS~\cite{yang2023deformable3dgs} and SCGS~\cite{huang2023sc}). We also provide our learned camera poses to baseline approaches for a fair comparison. These variants are denoted as ``w. Poses''. Best view in color and zoom in.}
    \label{fig:qualitative}
\end{figure}

\section{Experiment}
\label{sec:exp}
In this section, we provide an extensive evaluation of our method DGS with the initialization in Sec.~\ref{subsec:vidu4d}, comparing both appearance and geometry against previous state-of-the-art methods. Additionally, we analyze the contributions of each proposed component in detail.

\subsection{Implementation}
\label{subsec:Implementation}
For all qualitative and quantitative experiments, we follow the standard pipeline for dynamic reconstruction~\cite{DBLP:journals/tog/ParkSHBBGMS21}, to construct our evaluation setup by selecting every fourth frame as a training frame and designating the middle frame between each pair of training frames as a validation frame.

Our model configuration involves several key parameters to balance reconstruction and regularization losses. {For the field initialization stage,} we use a similar architecture with $8$ layers for volume rendering as in NeRF~\cite{DBLP:conf/eccv/MildenhallSTBRN20}, and initialize MLP for predicting SDF as an approximate unit sphere~\cite{DBLP:conf/nips/YarivKMGABL20}. We obtain a neural SDF, a warping field, and camera poses after this stage. {For the DGS stage,}
we initialize centers of the Gaussian surfels with the sampled surface points extracted from the neural SDF, and initialize the warping field by the forward field from the first stage. The dimension of the latent code embedding $\boldsymbol\gamma_b^t$ is set as $128$. Following BANMo~\cite{DBLP:conf/cvpr/YangVNRVJ22}, we adopt 25 bones to optimize skinning weights. For each reconstruction, the overall training takes over 1 hour on an A800 GPU.

\subsection{Qualitative Evaluation}
\label{subsec:quantitative}
In the qualitative evaluation, we visually compare the novel-view reconstructions produced by our DGS against those generated by other state-of-the-art models, as illustrated in Fig.~\ref{fig:qualitative}. Our evaluation focuses on several key aspects including detail preservation, texture quality, and geometric accuracy. Compared to methods based on implicit fields, the integration of Gaussian in our approach facilitates the rendering of highly detailed textures. Additionally, benefiting from a more geometry-aware representation, our method produces superior normal maps compared to those purely Gaussian-based methods. This also enhances the robustness of our method against artifacts of the generated videos like occlusions. For instance, in the third clip of the series, which features a dragon shrouded in fog, both SCGS and Deformable-GS methods tend to overfit and subsequently show a decline in performance. In contrast, our method consistently delivers superior results.

\begin{table}[t]
\centering
\caption{Novel-view quantitative results on generated videos. Evaluation metrics are PSNR, SSIM, and LPIPS. We report results on three single videos and the averaged results over 30 single videos.}
\tablestyle{3pt}{1.2}
\resizebox{1\columnwidth}{!}{
\begin{tabular}{@{}l|ccc|ccc|ccc|ccc@{}}
\toprule[1pt]
 & \multicolumn{3}{c@{}|}{Cat} & \multicolumn{3}{c@{}|}{Cheetah} & \multicolumn{3}{c@{}|}{Dragon} & \multicolumn{3}{c@{}}{Average over 30 videos}\\ 
& PSNR~$\uparrow$ & SSIM~$\uparrow$ & LPIPS~$\downarrow$ & PSNR~$\uparrow$ & 
SSIM~$\uparrow$ & LPIPS~$\downarrow$ & PSNR~$\uparrow$ & SSIM~$\uparrow$ & LPIPS~$\downarrow$ & PSNR~$\uparrow$ & SSIM~$\uparrow$ & LPIPS~$\downarrow$ \\
\midrule
BANMo~\cite{DBLP:conf/cvpr/YangVNRVJ22} & 15.10 & 0.6514 & 0.2575 & 13.15 & 0.5921 & 0.3241 & 18.48 & 0.6423 & 0.3500 & 13.62 $\pm$ 2.99 & 0.6153 $\pm$ 0.0714 & 0.3738 $\pm$ 0.0665 \\

D-NeRF~\cite{pumarola2021d} & 15.15 & 0.6537 & 0.2657 & 13.21 & 0.5930 & 0.3344 & 18.53 & 0.6489 & 0.3527 & 21.01 $\pm$ 2.86 & 0.8519 $\pm$ 0.0717 & 0.1522 $\pm$ 0.0754 \\

\midrule
Deformable-GS~\cite{yang2023deformable3dgs} & 19.09 & 0.7815 & 0.2434 & 20.35 & 0.8039 & 0.1982 & 24.19 & 0.9100 & 0.0992 & 13.22 $\pm$ 3.42 & 0.5934 $\pm$ 0.0535 & 0.3749 $\pm$ 0.0763 \\

SCGS~\cite{huang2023sc} & 19.46 & 0.7867 & 0.2405 & 20.87 & 0.8123 & 0.1919 & 24.03 & 0.9083 & 0.1009 & 21.17 $\pm$ 2.69 & 0.8547 $\pm$ 0.0691 & 0.1504 $\pm$ 0.0737 \\

Deformable-GS w. Poses & 21.94 & 0.8123 & 0.1816 & 22.41 & 0.8200 & 0.1687 & 26.05 & 0.9218 & 0.0894 & 22.63 $\pm$ 2.14 & 0.8469 $\pm$ 0.0438 & 0.1452 $\pm$ 0.0354 \\

SCGS w. Poses & 23.25 & 0.8268 & 0.1574 & 23.70 & 0.8338 & 0.1497 & 28.40 & 0.9375 & 0.0686 & 24.75 $\pm$ 2.11 & 0.8680 $\pm$ 0.0440 & 0.1201 $\pm$ 0.0359 \\

\textbf{DGS~(Ours)} & \textbf{24.63} & \textbf{0.8432} & \textbf{0.1559} & \textbf{25.68} & \textbf{0.8843} & \textbf{0.1117} & \textbf{28.58} & \textbf{0.9392} & \textbf{0.0618} & \textbf{27.30} $\pm$ \textbf{2.66} & \textbf{0.9152} $\pm$ \textbf{0.0602} & \textbf{0.0877}$\pm$ \textbf{0.0564} \\
\bottomrule[1pt]
\end{tabular}
}
\label{tab:main_results}
\end{table}

\subsection{Quantitative Evaluation}
\label{subsec:quantitative}
We provide the quantitative evaluation comparing our method with state-of-the-art works in Table~\ref{tab:main_results}. Metrics include Peak Signal-to-Noise Ratio (PSNR) to evaluate the fidelity of the reconstructed textures, Structural Similarity Index (SSIM) for the quality evaluation, and LPIPS~\cite{DBLP:conf/cvpr/ZhangIESW18} as a perceptual metric. Our method exhibits superiority over all baseline methods, even with our learned poses, \emph{e.g.}, $\sim$2.5 PSNR increase over SCGS with poses for the averaged results.

\begin{figure}[t]
    \centering
\includegraphics[width=1\columnwidth]{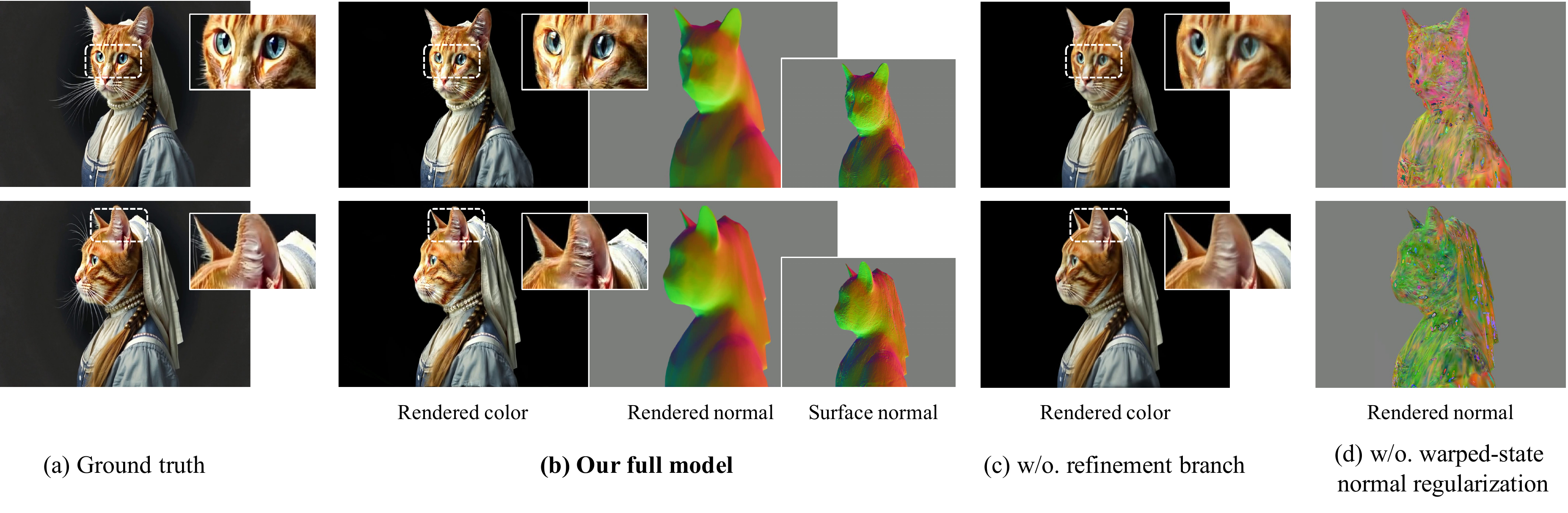}
    \vskip-0.1in
    \caption{{Ablation studies on the geometric regularization and refinement strategy.} For our full model shown in (b), we provide our rendered color, rendered normal, and surface normal (estimated from the depth points for regularization). Additionally, for comparison, we visualize the rendered color for the case without refinements in (c) and the rendered normal for the case without warped-state normal regularization in (d), respectively. We showcase our model's fidelity with close-ups.}
    \label{fig:ablation}
\end{figure}

\subsection{Ablations}
\label{subsec:ablation}
To understand the contributions of each component in Vidu4D, especially DGS, we conduct ablation studies in this section. We remove or alter specific elements of our model and observe the resulting performance changes in both appearance and geometry reconstruction.

\textbf{Geometric regularization.} We evaluate the impact of warped-state normal regularization by disabling it during training. From Fig.~\ref{fig:ablation}(b)(d), we observe that when removing the regularization, there is a significant degradation in the structural integrity of surface-aligned Gaussian surfels, leading to noticeable inconsistency in the reconstructed 4D models. 

\textbf{Refinement strategy.} We examine the effect of omitting refinements by keeping one branch (the concept of branches could be better visualized in Fig.~\ref{fig:method}) during training, shown in Fig.~\ref{fig:ablation}(b)(c). The performance indicates that removing refinements increases the loss of fine-grained appearance details. Additionally, we also find that refinements are crucial for mitigating the texture flickering issue. 

\textbf{Additional ablations.} Please refer to the Appendix for additional ablation studies that detail the effectiveness of our refinement strategy and field initialization.

\section{Conclusion}
\label{sec:conclusion}

We introduce Vidu4D as a novel reconstruction model to achieve high-fidelity 4D representations from single generated videos. Vidu4D is powerful with our proposed DGS which builds the non-rigid warping field to transform Gaussian surfels, ensuring precise capture of motion and deformation over time.  DGS also introduces key innovations that significantly enhance the accuracy and fidelity of 4D reconstruction, including dual branch refinement and warped-state geometric regularization. Our experiments demonstrate that Vidu4D outperforms existing methods in both quantitative and qualitative evaluations, highlighting its superiority in generating realistic and immersive 4D content. 

\textbf{Limitations and broader impact.} While Vidu4D with DGS presents a significant performance in 4D reconstruction, currently there are still limitations such as the reliance on video quality, scalability challenges for large scenes, and computational difficulties in real-time applications. Additionally, when equipping Vidu4D with generative models, as with any generative technology, there is a risk of producing deceptive content which needs more caution.

{\small
\bibliographystyle{splncs04.bst}
\bibliography{egbib}
}

\clearpage

\appendix

\section{Appendix / supplemental material}

\begin{table}[h]
\centering
\vskip -0.07in
\caption{\footnotesize{A summary of  important symbols  in DGS.}}
    \tablestyle{3pt}{1.2}
	\resizebox{1\linewidth}{!}{
		\begin{tabular}{ll}
			\toprule
			Symbol & Definition and Usage\\
			\midrule
			$\bt_u^*\in\mathbb{R}^{3\times 1}$, $\bt_v^*\in\mathbb{R}^{3\times 1}$&Principal tangential vectors in the static state.  \\
			$s_u^*\in\mathbb{R}$, $s_v^*\in\mathbb{R}$&Scaling factors in the static state.\\
			$\bp_k^*\in\mathbb{R}^{3\times 1}$& Center point coordinate (world space) of the $k$-th Gaussian surfel in the static state.\\
			$P_k^*(\bu)\in\mathbb{R}^{3\times 1}$& Coordinate (world space) in the static state, given  $\bu=(u,v)$ on the local $uv$ coordinate system centered at $\bp_k^*$.\\
			$\bR_k^*=[\bt_u^*,\bt_v^*, \bt_u^*\times \bt_v^*]\in\SO(3)$&  Rotation matrix of the $k$-th Gaussian surfel in the static state.\\
			$\bS_k^*=\mathrm{diag}(s_u^*, s_v^*, 0)\in\mathbb{R}^{3\times 3}$ & Scaling matrix of the $k$-th Gaussian surfel in the static state, a diagonal matrix. \\
			\midrule
			$\bp_k^t\in\mathbb{R}^{3\times 1}$& Center point coordinate (world space) of the $k$-th Gaussian surfel in the warped state.\\
			$P_k^t(\bu)\in\mathbb{R}^{3\times 1}$& Coordinate (world space) in the warped state, given  $\bu=(u,v)$ on the local $uv$ coordinate system centered at $\bp_k^t$.\\
			\midrule
			${\bc}^*_b\in\mathbb{R}^{3 \times 1}$, $\bV_b^*\in\mathbb{R}^{3\times3}$, $\boldsymbol{\Lambda}_b^*\in\mathbb{R}^{3\times3}$ & Center, rotation matrix, and diagonal scaling matrix of the $b$-th Gaussian ellipsoid bone. \\
			$\bw^t\in\mathbb{R}^{B\times1}$& Skinning weight vectors.\\
			$\bJ_b^{t}\in\SE(3)$ & A rigid transformation that moves the $b$-th bone from its static state to the warped state at time $t$.\\
			$\bJ^{t}=[
\tilde\bR^t,\tilde\bT^t]\in\SE(3)$ & The warping function, a weighted combination of $\bJ_b^{t}$.\\
			$\mathcal{Q}$, $\mathcal{R}$ & The quaternion process and the inverse quaternion process.\\
			\midrule
			$\boldsymbol\omega_b^t\in\mathbb{R}^{128}$& A learnable latent code for representing the body pose at time $t$.\\
			$\bn_k\in\mathbb{R}^{3\times 1}$ & The normal of the $k$-intersected Gaussian surfel that is oriented towards the camera.\\
			 $\bN^t\in\mathbb{R}^{3\times 1}$ & The surface normal estimated by the nearby depth point $\bp^t$ at warped state time $t$.\\
			 \midrule
			 $\Delta\bR_k^{*}\in\SO(3)$ & Learnable refinement term for  adjusting $\bR_k^{*}$.\\
			 $\Delta\bS_k^{*}\in\SO(3)$ & Learnable refinement term for  adjusting $\bS_k^{*}$.\\
			\bottomrule
		\end{tabular}
	}
\label{table:symbol}
\end{table}

\subsection{Ablation: Field Initialization and Refinement}
\label{subsec:field init}
\begin{figure}[t]
    \centering
    \includegraphics[width=1\columnwidth]{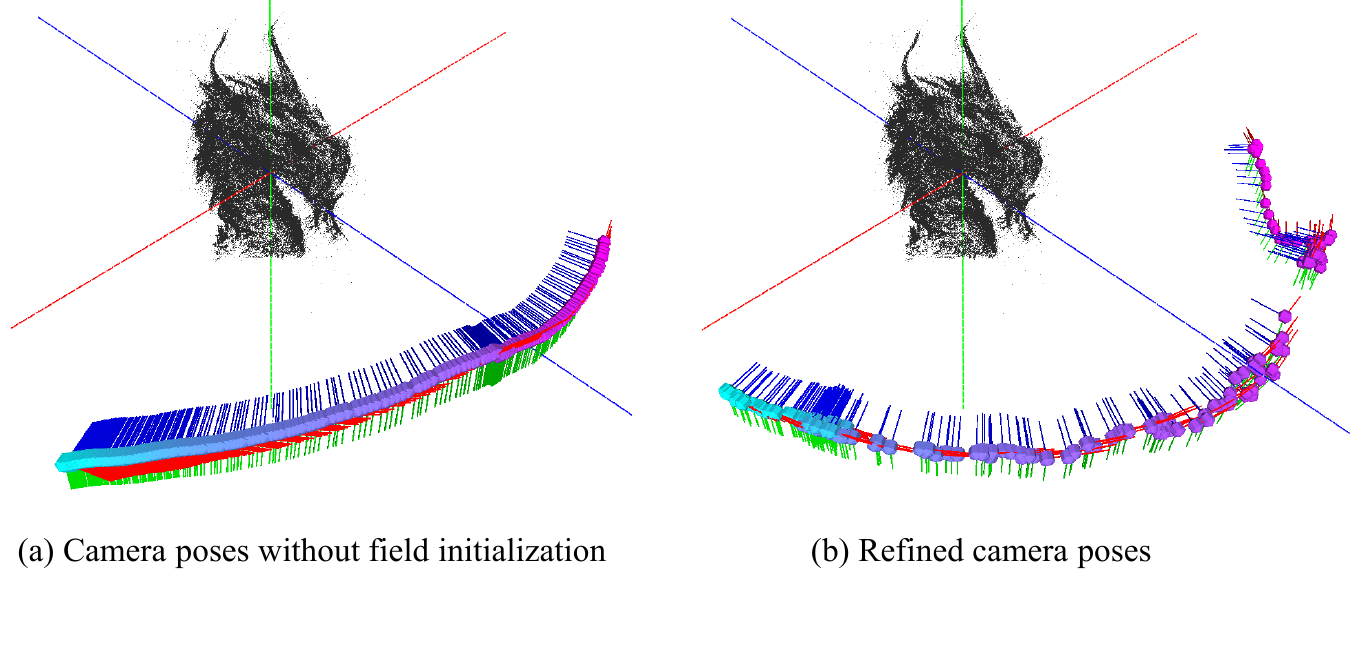}
    \vspace{-1 cm}
    \caption{Coarse camera poses and refined camera poses.}
    \label{fig:supp-cam}
\end{figure}

In dynamic videos captured in the wild, one of the primary challenges is the initialization of camera poses. In synthetic videos, preserving temporal consistency in texture and geometry is problematic, which significantly complicates the task of camera registration. To address this, we utilize an implicit field to both initialize the camera poses and establish the warping field. Initially, we estimate the transformation for each frame, followed by the computation of coarse camera poses through an iterative process. Subsequently, we adopt the approach outlined in NeuS~\cite{DBLP:conf/nips/WangLLTKW21} for scene representation. Feature extraction is performed using DinoV2~\cite{oquab2023dinov2}, facilitating unsupervised registration. To enhance this process, we train an additional channel in NeuS specifically for rendering features, which are then employed for registration purposes as described in RAC~\cite{yang2023reconstructing}. The camera poses without initialization and refined camera poses are depicted in Fig.~\ref{fig:supp-cam}. Without field initialization, the performance of DGS will degrade, as shown in Table~\ref{tab:supp-abl}. Also, please refer to the quantitative ablation of refinement in Table~\ref{tab:supp-abl}.

\begin{table}[h!]
\centering
\caption{Quantitative ablation studies of the initialization and refinement.}
\tablestyle{3pt}{1.2}
\resizebox{1\columnwidth}{!}{
\begin{tabular}{@{}l|ccc|ccc|ccc@{}}
\toprule[1pt]
 & \multicolumn{3}{c@{}|}{Cat} & \multicolumn{3}{c@{}|}{Cheetah} & \multicolumn{3}{c@{}}{Dragon}\\ 
& PSNR~$\uparrow$ & SSIM~$\uparrow$ & LPIPS~$\downarrow$ & PSNR~$\uparrow$ & 
SSIM~$\uparrow$ & LPIPS~$\downarrow$ & PSNR~$\uparrow$ & SSIM~$\uparrow$ & LPIPS~$\downarrow$ \\
\midrule
Ours w.o. init.  & 20.15 & 0.7961 & 0.2393 & 20.96 & 0.8194 & 0.1940 & 25.33 & 0.9146 & 0.0938 \\
Ours w.o. refinement  & 24.19 & 0.8196 & 0.1797 & 24.10 & 0.8582 & 0.1242 & 27.71 & 0.9128 & 0.0687 \\
\textbf{Ours full} & \textbf{24.63} & \textbf{0.8432} & \textbf{0.1559} & \textbf{25.68} & \textbf{0.8843} & \textbf{0.1117} & \textbf{28.58} & \textbf{0.9392} & \textbf{0.0618} \\
\bottomrule[1pt]
\end{tabular}
}
\label{tab:supp-abl}
\end{table}

\subsection{Additional Qualitative Comparison}
\label{subsec:field init}
\begin{figure}[t]
    \centering
    \includegraphics[width=1\columnwidth]{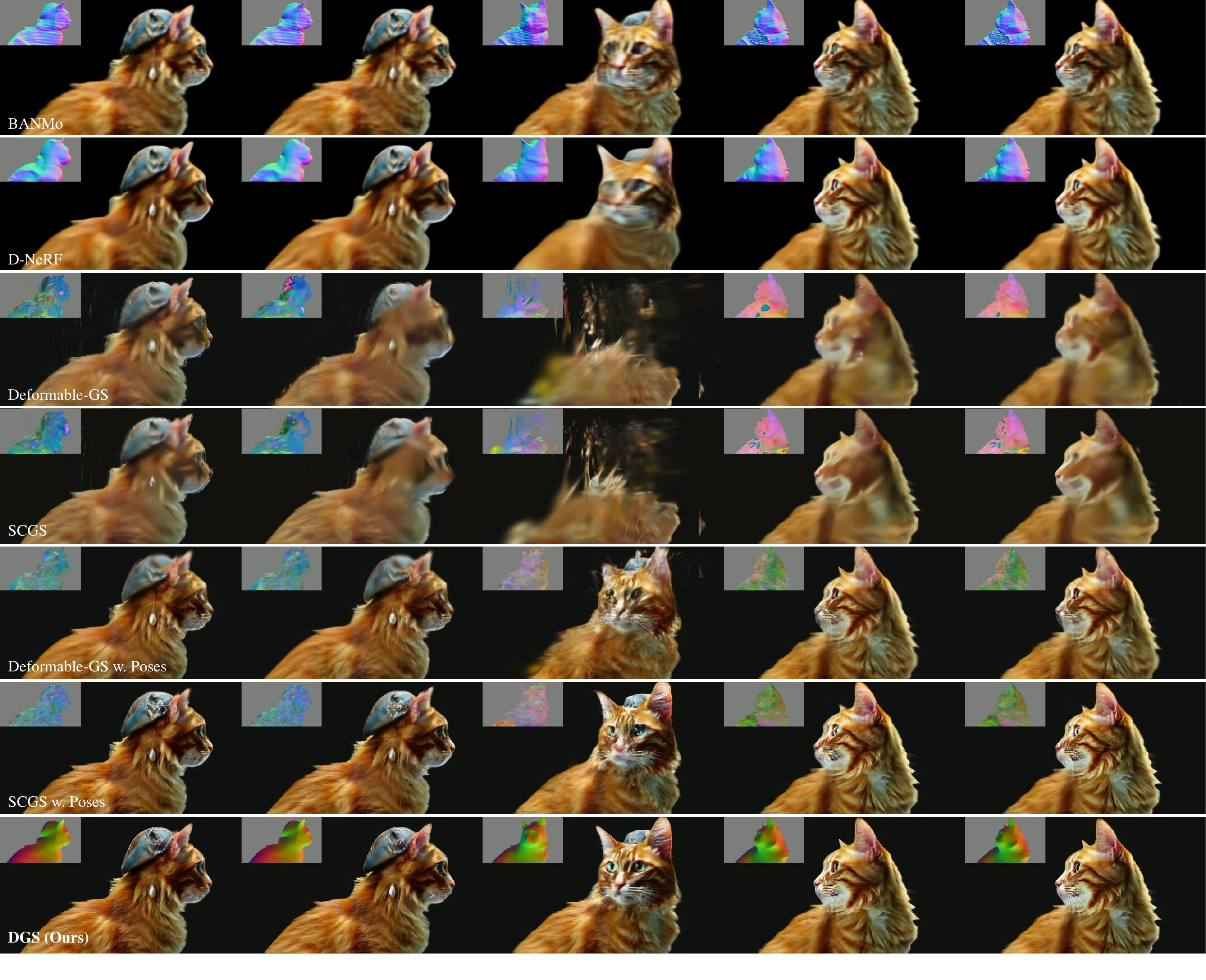}
    \caption{Additional qualitative comparison with more novel views.}
    \label{fig:supp-2}
\end{figure}

\begin{figure}[t]
    \centering
    \includegraphics[width=1\columnwidth]{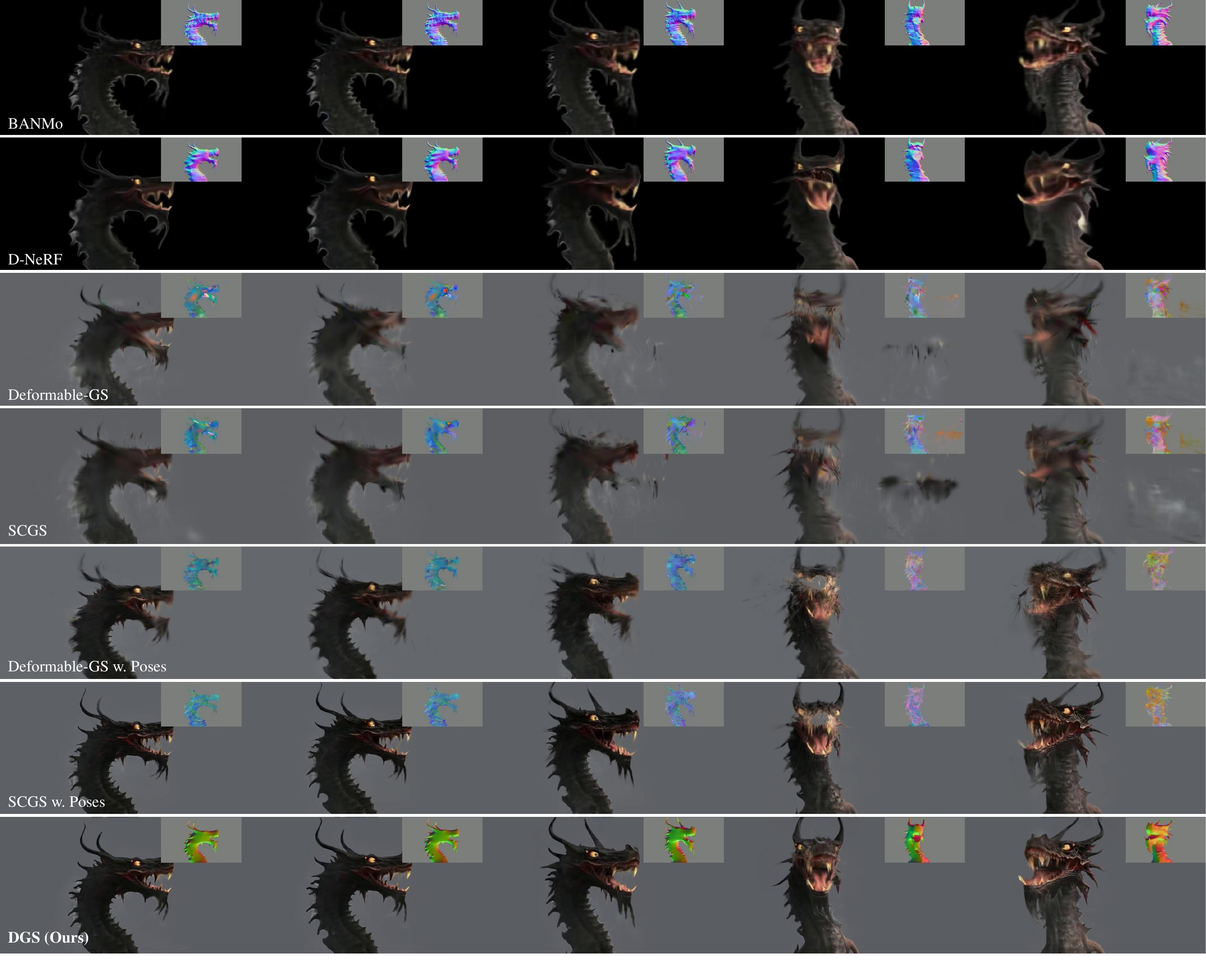}
    \caption{Additional qualitative comparison with more novel views.}
    \label{fig:supp-3}
\end{figure}

\begin{figure}[t]
    \centering
    \includegraphics[width=1\columnwidth]{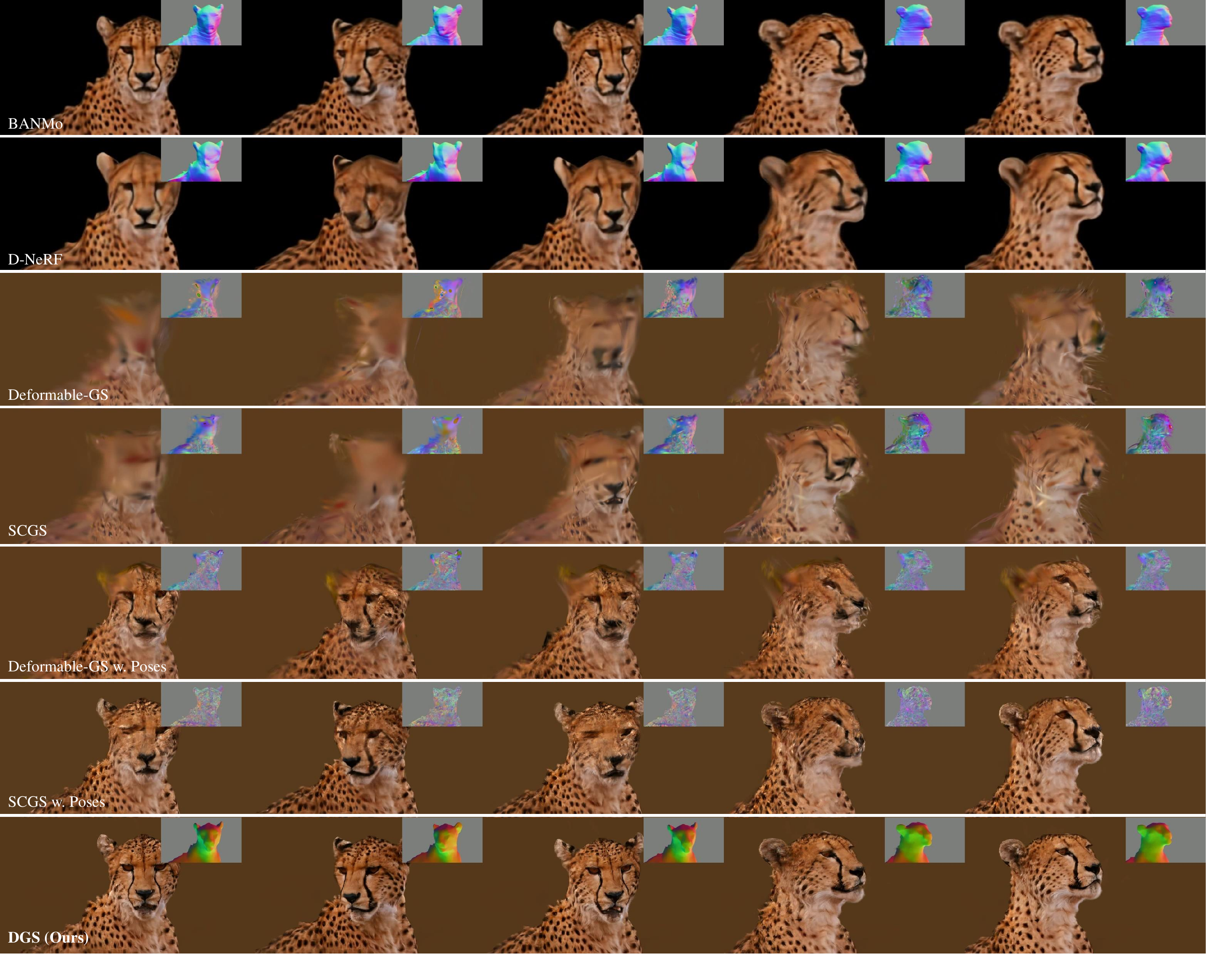}
    \caption{Additional qualitative comparison with more novel views.}
    \label{fig:supp-4}
\end{figure}

\begin{figure}[t]
    \centering
    \includegraphics[width=1\columnwidth]{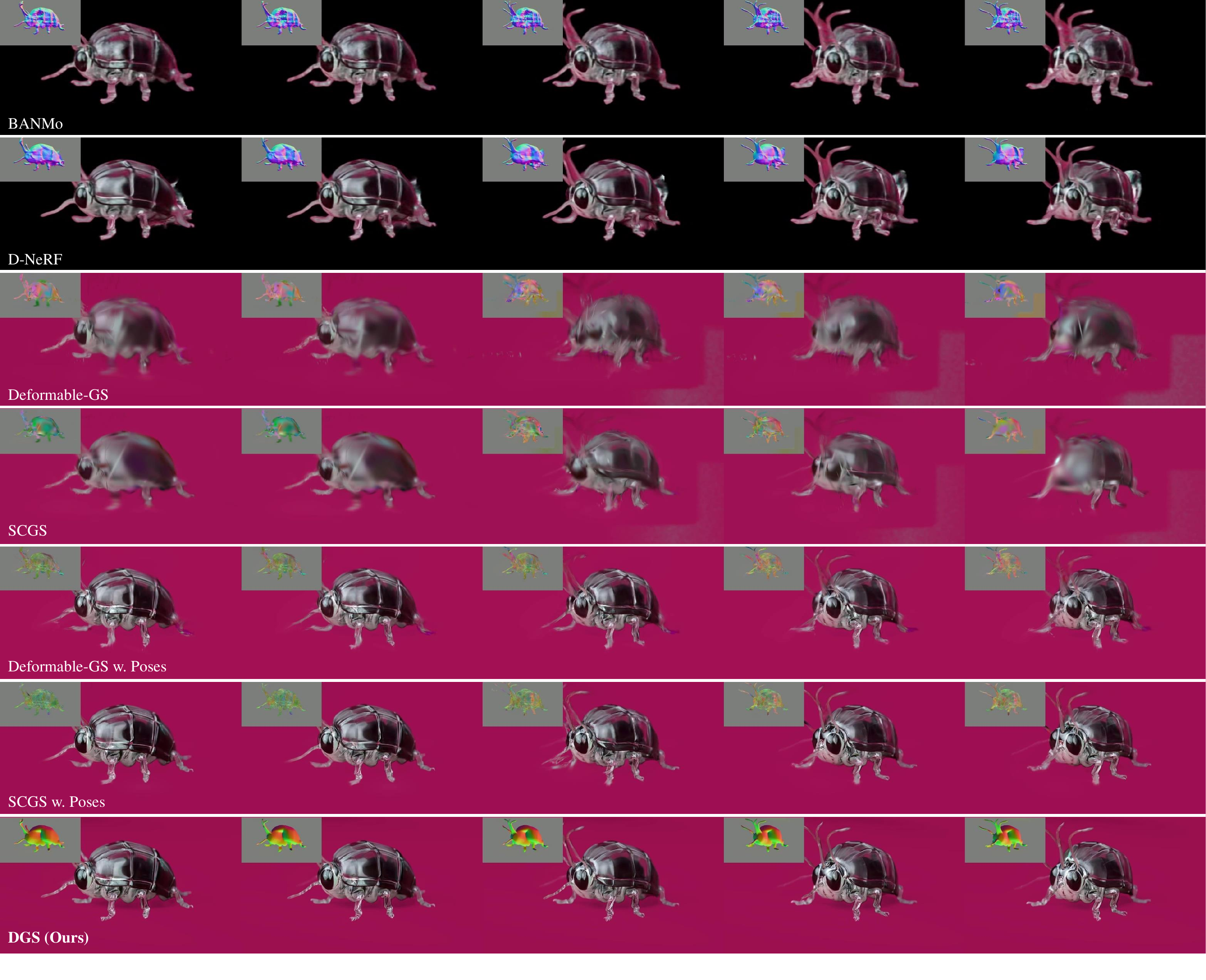}
    \caption{Additional qualitative comparison with more novel views.}
    \label{fig:supp-5}
\end{figure}

In this section, we present a detailed comparison of our results with previous works, as illustrated in Fig.~\ref{fig:supp-2}-\ref{fig:supp-5}. Our method consistently achieves high-quality texture details while maintaining smooth and realistic geometry.

\subsection{Interpolation on Time and Views}
\label{subsec:field init}
\begin{figure}[t]
    \centering
    \includegraphics[width=1\columnwidth]{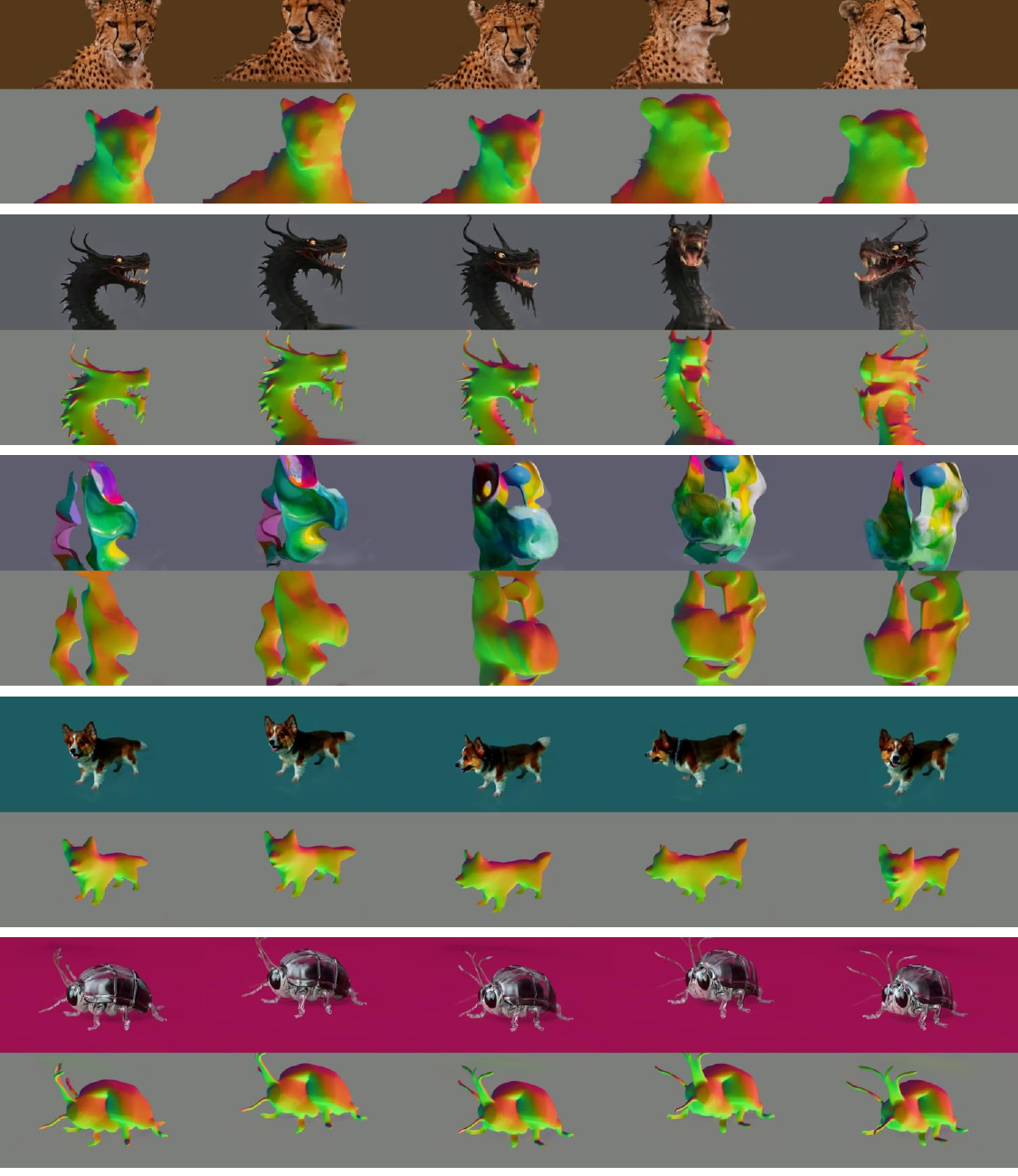}
    \caption{Interpolation on time and views.}
    \label{fig:supp-6}
\end{figure}

\begin{figure}[t]
    \centering
    \includegraphics[width=1\columnwidth]{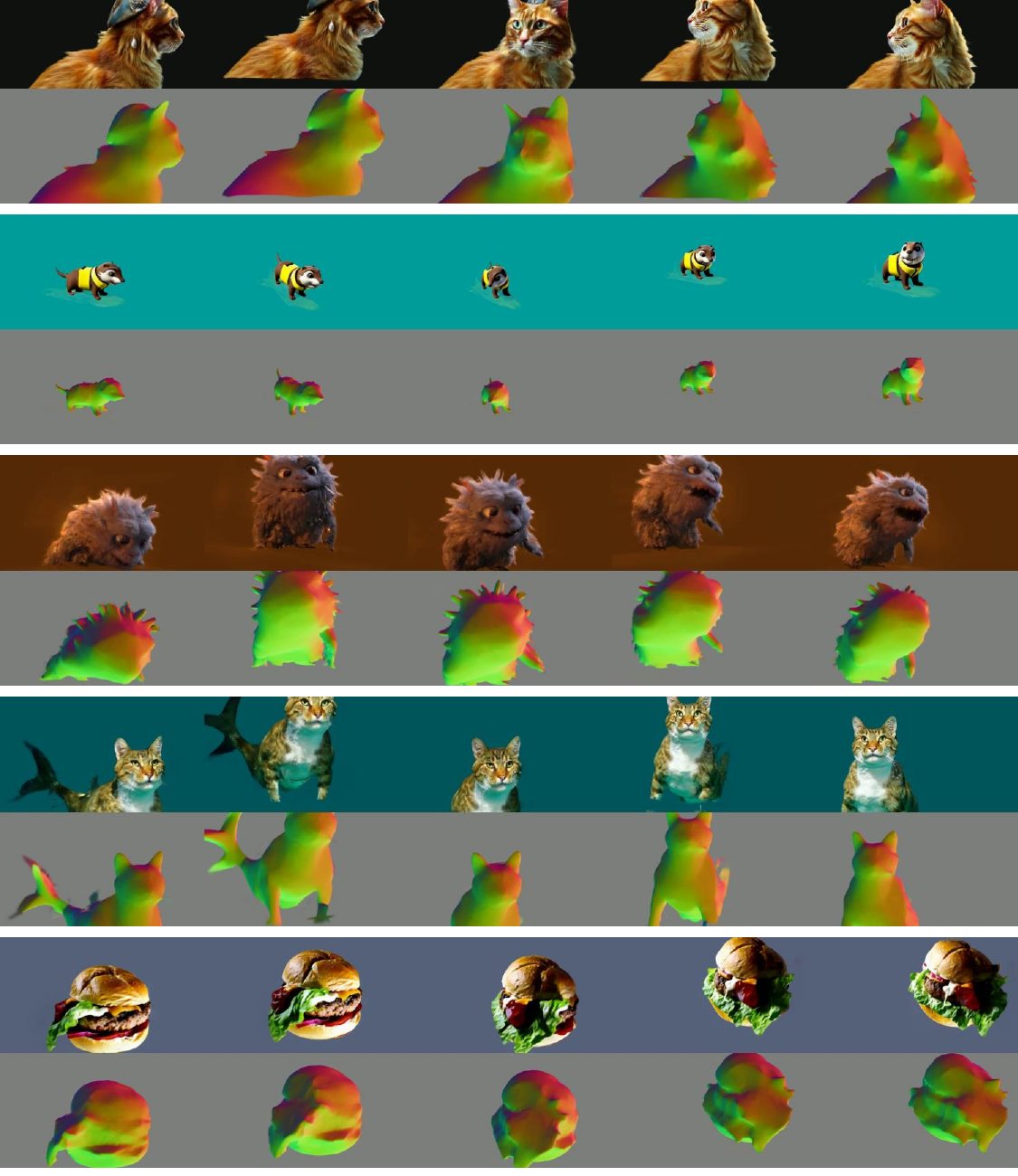}
    \caption{Interpolation on time and views.}
    \label{fig:supp-7}
\end{figure}

We present results for interpolation on time and views, as illustrated in Fig.~\ref{fig:supp-6} and Fig.~\ref{fig:supp-7}. 

\end{document}

%% file: math_commands.tex
\usepackage{amsmath,amsfonts,bm}

\def\1{\bm{1}}

\def\vg{{\bm{g}}}

\DeclareMathAlphabet{\mathsfit}{\encodingdefault}{\sfdefault}{m}{sl}
\SetMathAlphabet{\mathsfit}{bold}{\encodingdefault}{\sfdefault}{bx}{n}

\newcommand{\bc}{\mathbf{c}}

\newcommand{\bJ}{\mathbf{J}}

\newcommand{\bn}{\mathbf{n}}\newcommand{\bN}{\mathbf{N}}

\newcommand{\bp}{\mathbf{p}}

\newcommand{\bR}{\mathbf{R}}
\newcommand{\bS}{\mathbf{S}}
\newcommand{\bt}{\mathbf{t}}\newcommand{\bT}{\mathbf{T}}
\newcommand{\bu}{\mathbf{u}}
\newcommand{\bV}{\mathbf{V}}
\newcommand{\bw}{\mathbf{w}}
\newcommand{\bx}{\mathbf{x}}

\newcommand{\cG}{\mathcal{G}}

\DeclarePairedDelimiterX{\infdivx}[2]{(}{)}{%
    #1\;\delimsize\|\;#2%
}